\documentclass{article}
\usepackage{amsmath}

\usepackage{PRIMEarxiv}
\usepackage[utf8]{inputenc} 
\usepackage[T1]{fontenc}    
\usepackage{hyperref}       
\usepackage{url}            
\usepackage{booktabs}       
\usepackage{amsfonts}       
\usepackage{nicefrac}       
\usepackage{microtype}      
\usepackage{lipsum}
\usepackage{fancyhdr}       
\usepackage{graphicx}       
\graphicspath{{media/}}     
\usepackage{subfigure}
\title{SCRNet: a Retinex Structure-based Low-light Enhancement Model Guided by Spatial Consistency}

\author{
  Miao Zhang \\
  \\
  \texttt{zhangmiaotju@gmail.com} \\
  \And
  Yiqing Shen \\
  Johns Hopkins University \\
  \texttt{yshen92@jhu.edu} \\
   \And
  Shenghui Zhong \\
  Beihang Hangzhou Innovation Institute Yuhang \\
  \texttt{zhongshenghui@buaa.edu.cn} \\
}

\begin{document}
\maketitle

\begin{abstract}
    Images captured under low-light conditions are often plagued by several challenges, including diminished contrast, increased noise, loss of fine details, and unnatural color reproduction. These factors can significantly hinder the performance of computer vision tasks such as object detection and image segmentation. As a result, improving the quality of low-light images is of paramount importance for practical applications in the computer vision domain.
  To effectively address these challenges, we present a novel low-light image enhancement model, termed Spatial Consistency Retinex Network (SCRNet), which leverages the Retinex-based structure and is guided by the principle of spatial consistency. 
  Specifically, our proposed model incorporates three levels of consistency: channel level, semantic level, and texture level, inspired by the principle of spatial consistency. 
  These levels of consistency enable our model to adaptively enhance image features, ensuring more accurate and visually pleasing results. 
  Extensive experimental evaluations on various low-light image datasets demonstrate that our proposed SCRNet outshines existing state-of-the-art methods, highlighting the potential of SCRNet as an effective solution for enhancing low-light images.
\end{abstract}


\section{Introduction}
\label{sec:intro}

Computer vision algorithms are heavily reliant on high-quality visible input images, as highlighted by research studies \cite{MOVINGDARK,YOLODARK,HLAFACE}. However, capturing images under low-light conditions often poses several challenges, such as a decrease in contrast, the presence of unexpected noise, loss of fine details, and unnatural color reproduction. These degradation issues can significantly increase the complexity of high-level tasks such as object detection and image segmentation, making it challenging to obtain accurate results. Therefore, enhancing low-light images is of immense practical value in the field of computer vision.

Various adjustments can be made to the camera imaging parameters to acquire high-quality images in low-light environments, such as increasing the ISO, extending the exposure time, or utilizing a flash. However, each of these methods has its limitations. For example, a high ISO setting can amplify the image sensor's sensitivity to light, but it can also intensify noise, leading to a low signal-to-noise ratio (SNR). Prolonged exposures can result in blurred outcomes, especially when capturing dynamic scenes, and using a flash can cause unbalanced lighting and unnatural colors in the photograph. As a result, developing an effective low-light image enhancement technique that can simultaneously reduce darkness and mitigate degradation issues is of utmost importance. Overcoming these challenges remains a significant obstacle in the field of computer vision.
\begin{figure}
\centering
\subfigure[Low-light image]
{\label{Low-light image}
\includegraphics[width=0.55\textwidth]{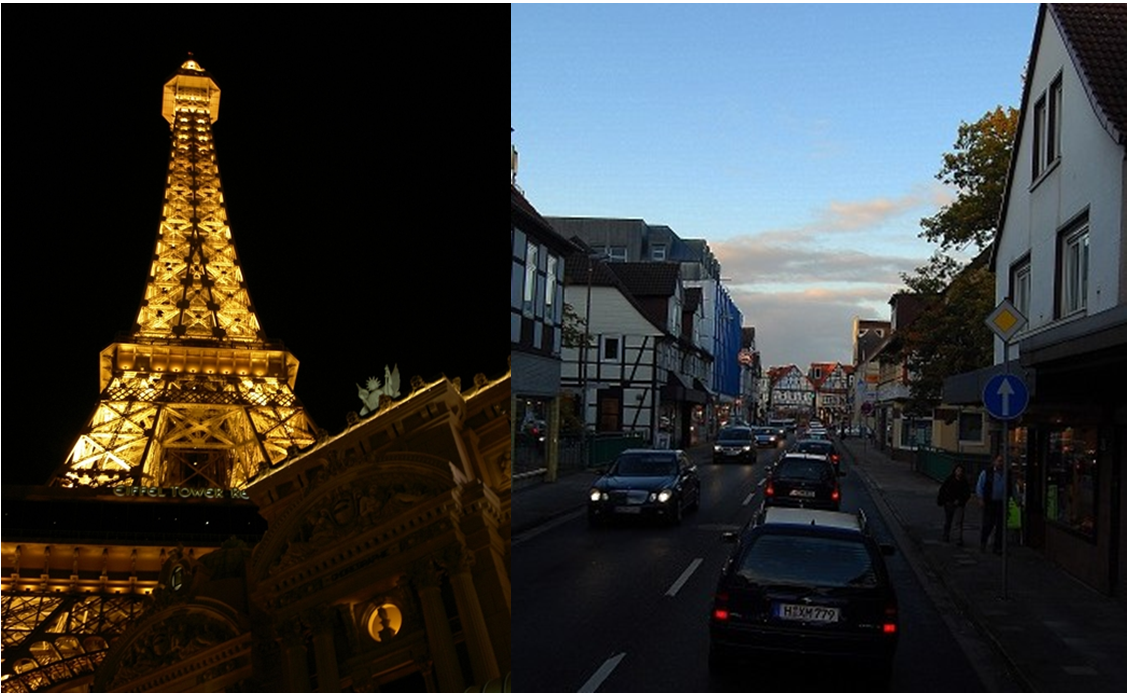}}
\subfigure[Enhanced image]{\label{Enhanced image}     \includegraphics[width=0.55\textwidth]{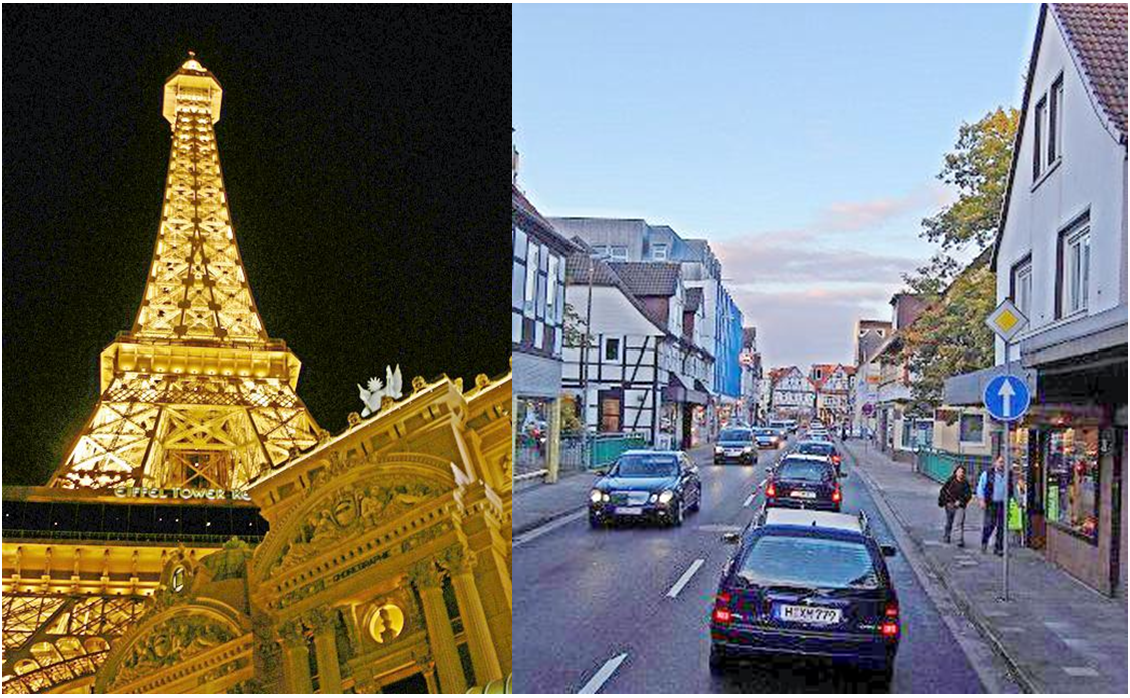}}
\caption{Example of enhancement process for low-light images}
\label{fig:intro} 
\end{figure}

To address the challenges of enhancing low-light images, a range of algorithms have been proposed to improve the subjective and objective quality of such images, as illustrated in Figure \ref{fig:intro}. These methods can be classified into three categories: distribution mapping-based methods, Retinex-based methods, and deep learning-based methods.
Distribution mapping-based methods utilize techniques like curve transformation and histogram equalization to alter pixel distribution, resulting in clearer and more emphasized images \cite{HISTO-1, HISTO-2}. However, these methods lack the ability to extract intrinsic distribution and semantic information between pixels, leading to color distortion and detail anomalies in the enhanced images \cite{HISTO-3}.
Retinex-based approaches are grounded in the principle that images can be decomposed into reflection and illumination components \cite{retinex-1,retinex-2}. Reflection constitutes an inherent property of a scene, while illumination is influenced by ambient light conditions. However, conventional Retinex algorithms often result in underexposed images, unsaturated colors, and significant artifacts or noise due to their design constraints and reliance on a priori knowledge \cite{retinex-3,retinex-4}.
In recent years, learning-based methods \cite{retinex-d1,retinex-d2,retinex-d3} have emerged as the mainstream approach for enhancing low-light images, with rapid advancements in deep learning technology. These methods establish mappings between low-light inputs and enhanced images by designing heuristic network structures that can adaptively learn complex relationships between input and output images, resulting in accurate and visually appealing enhancements.

Although learning-based models have shown promising results in enhancing low-light images, they often process RGB images in their entirety, ignoring the distinct detail information carried by each color channel. This oversight can lead to interference between channels and subsequent loss of details. To address this issue, we propose a novel enhancement and denoising model for R, G, and B channels on the assumption of single-channel consistency.

Our model builds upon the Retinex model and comprises two primary components: Decomposition and Enhancement. The Decomposition component accepts a low-light image as input and generates the illumination map and reflectance map through semantic, instance, and texture pathways. The Enhancement component employs the reflectance map for detail restoration, denoising, and progressive color correction, adhering to semantic and texture consistency principles. The brightness and contrast of the illumination map are adjusted using the Adjustment network. Ultimately, the corrected illumination map and the reflectance map are combined to produce the enhanced image. Empirical results indicate that our method generates accurate illumination maps and achieves more natural results with superior details compared to state-of-the-art techniques, both qualitatively and quantitatively.

This work's primary contributions can be summarized as follows:
\begin{itemize}
    \item 
    We introduce the Cascading Texture-Instance-Semantic Feature Fusion Module (CFM), which comprises three branches that capture semantic, instance, and texture information separately. Secondly, we propose the Channel-dependent Spatially Consistent Denoising Module (CDM), which performs bilateral filtering on each of the three RGB channels individually to achieve spatially consistent denoising.
\end{itemize}
\begin{itemize}
    \item For the task of Detail Restoration, we present an encoder-decoder structure and introduce a novel and effective Regional Consistency-based Non-rigid Sampling Pyramid Module (RPM), which concentrates on restoring the dark regions of the image based on the illumination distribution at different scales.
\end{itemize}
\begin{itemize}
    \item We introduce the Progressive Binomial Color Correction Module (PCM) in this paper, which utilizes a color matrix with binomial expansion for more effective nonlinear color correction.
\end{itemize}
\begin{itemize}
    \item We conducted comprehensive experiments on several benchmark datasets to demonstrate the superiority of our method over existing state-of-the-art methods. Additionally, we performed an ablation study to validate the effectiveness of our proposed structure. 
\end{itemize}

\begin{figure*}[htp]
    \centering
    \includegraphics[scale=0.60]{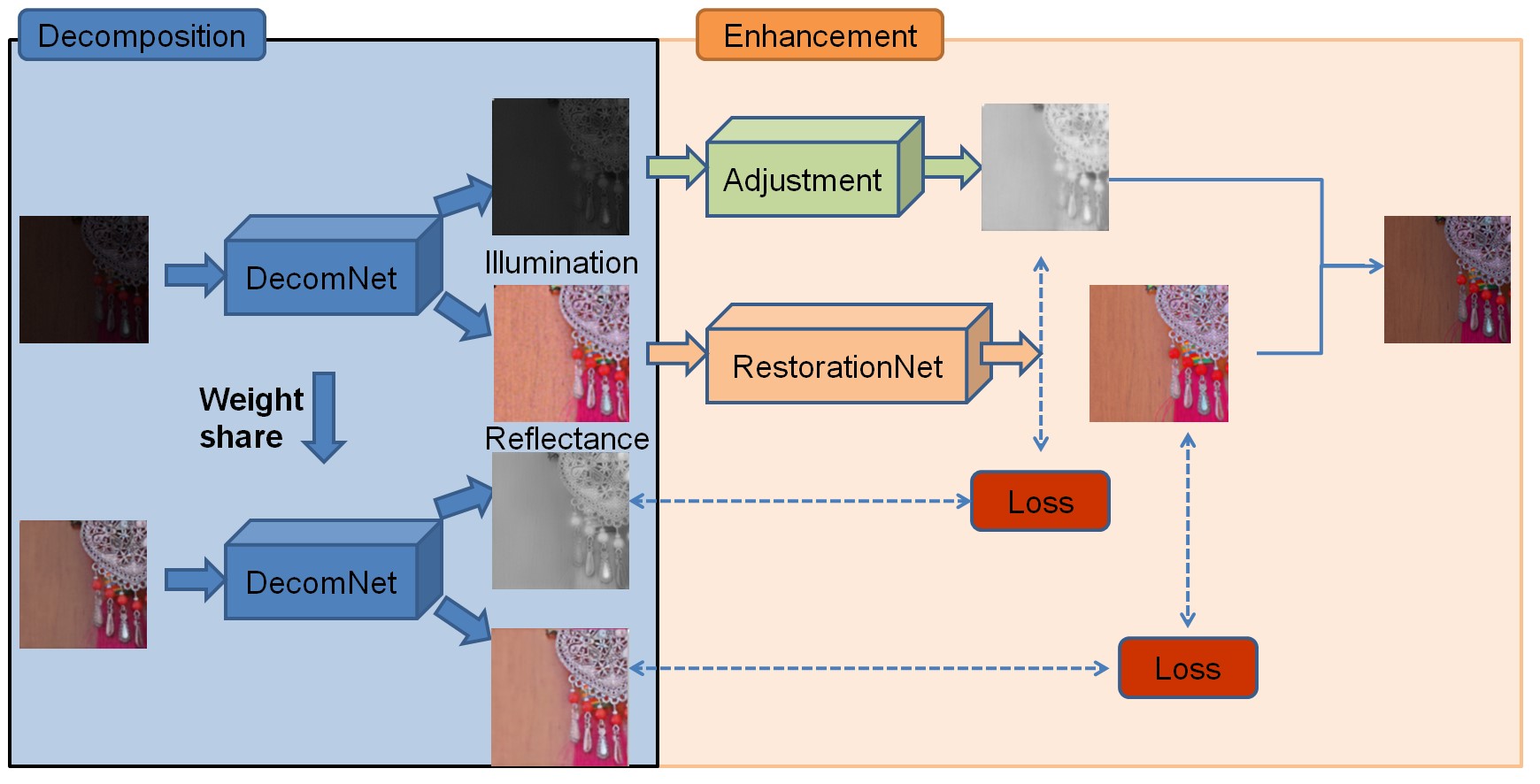}
    \caption{The overview of our method The overview of the proposed SCRNet. That consists of the decomposition part and enhancement part}
    \label{fig:denet}
\end{figure*}
\section{Related Works}
\subsection{Learning-based enhancement methods.}

Li et al. \cite{LLNet} proposed LLNet, a method that utilized two distinct deep self-encoder networks that operated independently. One network acted as an encoder, transforming the low-light image into a low-dimensional representation, while the other network functioned as a decoder, converting the low-dimensional representation back into the original image space. By employing these networks, the method could autonomously learn signal characteristics and noise structures present in the image. Additionally, they introduced two loss functions, luminance loss and structural similarity loss, to optimize the network during training. LLNet demonstrated significant improvements in enhancing low-light images.
Wei et al. \cite{retinex} proposed Deep Retinex Decomposition, a method that deviated from traditional Retinex methods that typically depend on manually set parameters for decomposition. Instead, they utilized a deep neural network to decompose the image into more accurate components, effectively enhancing the brightness and contrast of the image while preserving its details and color information. The proposed method employed a data-driven approach to learn the decomposition parameters, making it more adaptive to various image characteristics.

Building on the RetinexNet architecture from the previous paper, Zhang et al. \cite{kind, kind_plus} introduced additional training loss constraints and considered a range of image enhancement metrics such as brightness, contrast, and structure in the loss design. This approach enabled a more comprehensive evaluation of the image enhancement effect.
As research progressed, Li et al. \cite{pyramid} proposed a lightweight and efficient model structure for joint brightness enhancement and noise removal. They introduced a coarse-to-fine architecture and luminance-aware strategy to design a lightweight and efficient block called MSCFB (Multi-Scale Concatenated Feature Block) as a crucial component of the network. This block could simultaneously extract image features at different scales and exploit contrast information to obtain high-quality images with natural colors and rich details. By integrating brightness enhancement and noise removal into a single network, their method outperformed previous methods that addressed these tasks separately.

\subsection{Learning-based Joint Brightness Enhancement and Noise Removal}
In some complex scenes, the image enhancement process can introduce or generate noise and artifacts, which can diminish image quality. To tackle this issue, certain methods aim to achieve simultaneous luminance enhancement and noise removal to generate high-quality images.
For example, Gharbi et al. \cite{bilateral} proposed a deep learning-based bilateral filtering algorithm for real-time image enhancement. Their algorithm predicted the coefficients of local affine models in bilateral space using deep learning, combining the advantages of traditional bilateral filtering with the ability to preserve image details while removing noise and blur.
Jiang et al. \cite{DEANet} leveraged prior knowledge that noise primarily resides in the high-frequency information of images. They proposed DEANet (Dual-branch Enhanced Attention Network), a method that used a weighted least squares (WLS) filter to separate low-light images into high-frequency images, which store noise and boundary information, and low-frequency images, which store detail and luminance. By separating these image components, their method achieved superior image enhancement and denoising. Furthermore, they introduced a dual-branch enhanced attention mechanism that adaptively weighted the contribution of each component to the final output, resulting in improved image quality.
    
\section{Methods}
In this section, we first provide an overview of the Retinex model, which serves as the foundation for our approach. We then introduce the novel components proposed in our model and explain how we estimate the illumination and reflectance (as well as noise), including details on the network structure, loss function, and implementation.
\subsection{Retinex Model}
Our network is based on the Retinex model, which has been widely used for low-illumination image enhancement and describes the luminance and color perception of human vision. 
In the Retinex model, the observed image (i.e., weakly illuminated image) is represented by \textit{S}, the reflectance component is represented by \textit{R}, the illumination component is represented by \textit{I}, and noise is represented by \textit{N}. 
The model describes the relationship between these components as shown in Eq. \ref{eq:1}, where $\{r, g, b\}$ represents the RGB channels, and $*$ represents the pixel-level multiplication. 
Since different color channels (i.e., red, green, and blue) may have distinct noise characteristics, we perform denoising and color correction separately on each channel. 
This ensures that the enhancement process is performed consistently across all channels.
\begin{equation}
S=\left(R^{*} I+N\right)_{\{r, g, b\}}\label{eq:1}
\end{equation}

\subsection{Overall Network Architecture}
An effective low-light image enhancement model must recover image details while also addressing the challenges of noise, color distortion, and degradation that frequently occur in low-light environments. To achieve this, we propose a new deep network architecture consisting of two main components: a Decomposition component and an Enhancement component, as illustrated in Figure \ref{fig:denet}. The Decomposition component includes a decomposition network that separates the low-light image into its illumination and reflectance components while reducing noise. The Enhancement component comprises two separate networks: a restoration network that focuses on improving the quality of the reflectance component, and an adjustment network that enhances contrast and adjusts the lighting of the illumination component. In the following sections, we will describe these networks in detail, explaining how they function and contribute to the overall effectiveness of our model in enhancing low-light images.
\begin{figure*}[htp]
    \centering
    \includegraphics[scale=0.6]{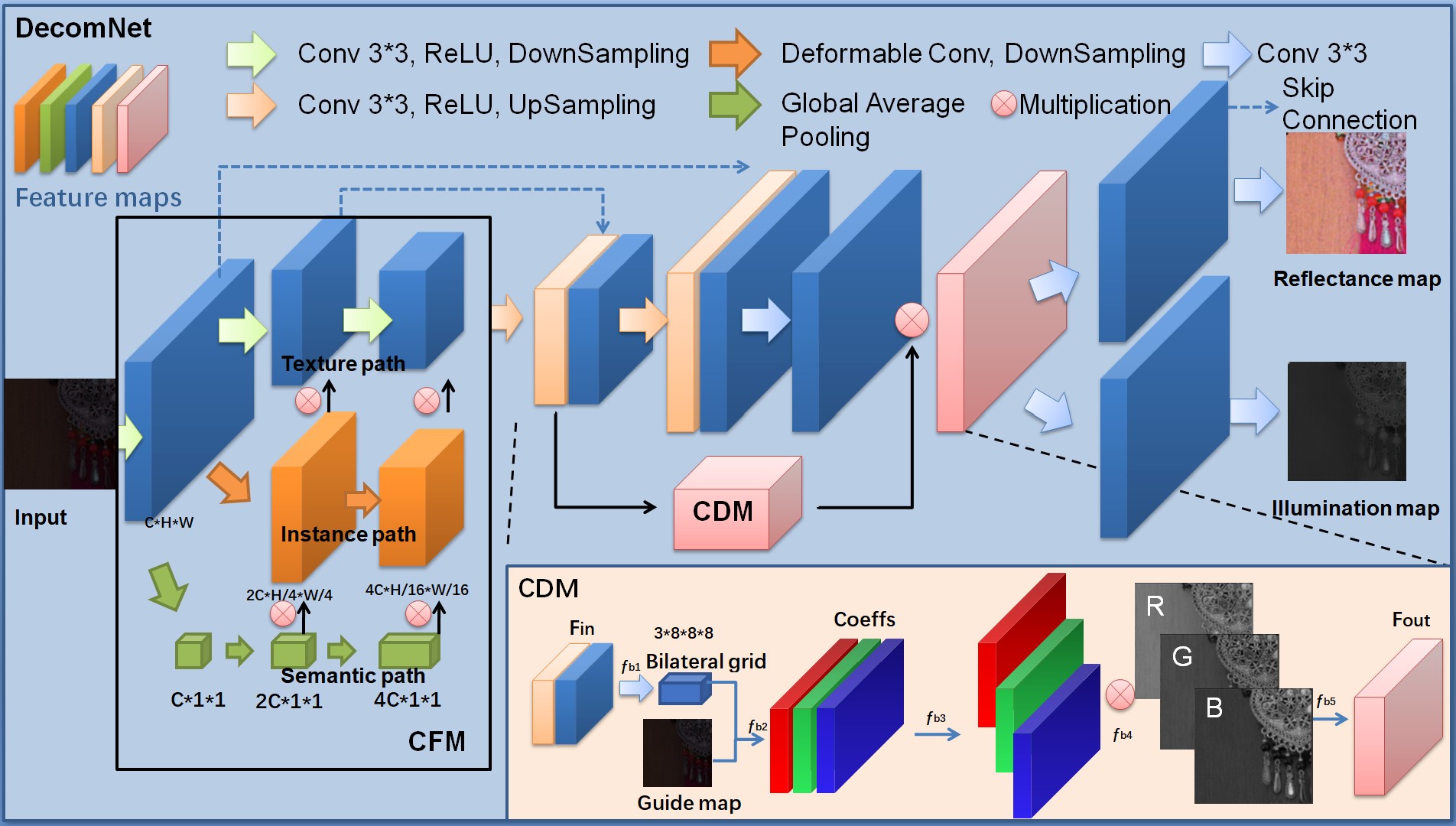}
    \caption{The architecture of Decomposition. The encoder-decoder Net mainly includes Cascading Texture-Instance-Semantic Feature Fusion Module (CFM)  and Channel-dependent Spatially Consistent Denoising Modules (CDM).}
    \label{fig:decomnet}
\end{figure*}

\subsection{Decomposition}
The Decomposition component, as illustrated in Figure \ref{fig:decomnet}, takes the input image and decomposes it into two components: \textit{R} and \textit{I}. This is achieved by passing the input image through an encoder-decoder CNN network that extracts feature maps. The network employs a two-path approach: one path incorporates semantic information inspired by \cite{SENet} to obtain high-level features, while the other path uses a U-Net-like structure to obtain instance and texture features at lower levels. These features are then fused together to obtain a combined feature map. Finally, bilateral filtering is applied to denoise and correct the R-, G-, and B-channels individually, ensuring that the enhancement process is performed consistently across all color channels.

\paragraph{Cascading Texture-Instance-Semantic Feature Fusion Module (CFM)}
Enhancing under-exposed photos is a challenging task that requires adjustments to texture, instance, and semantic features. Semantic features refer to the overall properties of the image, such as color balance, brightness, and scene category, while texture features pertain to specific regions of the image and include elements such as highlighting detail, sharpness, and contrast. Traditional image enhancement algorithms often focus on adjusting either texture features, such as methods based on local contrast enhancement \cite{salih2018localcontrast}, or on adjusting semantic features, such as histogram equalization \cite{reddy2022enhancement}. However, balancing both aspects can be difficult \cite{huang2021global}. Some deep-learning-based methods extract semantic and texture features only on a single feature map \cite{wang2019underexposed, bilateral}, or only fuse them once at the end of the model \cite{cui2022IAT}. To address these limitations, we introduce CFM (Cascading semantic, instance, and texture feature fusion module), which characterizes semantic, instance, and texture features from different scale feature maps and fuses them at each block. This allows our model to capture and leverage the complementary information in these distinct features, leading to more effective and balanced image enhancement.

To capture semantic features (Semantic path), we employ a global average pooling operation to compress the feature map \textit{Fin} with dimensions \textit{C*H*W} into a vector of size \textit{C*1*1}. We then use a fully connected (FC) operation for channel expansion to expand the number of channels to 2C, enabling the representation of different channel combinations. Finally, the channel size is expanded to \textit{4*C*1*1} to represent the importance of each channel. This is achieved by multiplying the weights by the original feature map \textit{Fin}, producing a more robust feature representation, \textit{Fout}.

To capture instance features (Instance path), we use a deformable convolutional neural network (CNN) that can capture irregular instance information. The feature map with dimensions \textit{C*H*W} is continuously downsampled to \textit{2*C*H/4*W/4} and \textit{4*C*H/16*W/16} to extract instance features at different scales.
For texture features (Texture path), we utilize a classical encoder-decoder structure to extract texture features. To avoid gradient vanishing and information loss during the upsampling process, we employ a skip-connection structure where the output of each convolutional layer is connected to the input of the corresponding deconvolutional layer. This allows the model to learn detailed information in the image while also preserving low-level information.

\paragraph{Channel-dependent Spatially Consistent Denoising Modules (CDM)}

When forming an image, noise is unavoidably introduced, such as image grain noise and color noise. Previous studies \cite{hai2023r2rnet, bilateral, DEANet} have applied global denoising methods on all three RGB channels, which assume that the information is uniformly distributed across the various channels. However, this assumption is frequently not valid, and this technique may lead to various types of noise being treated as identical noise types. As a result, inter-channel interference occurs, diminishing the efficacy of denoising.

To avoid inter-channel interference and improve the denoising effect, we propose a separate denoising approach for each of the three RGB channels, coupled with bilateral filtering. Unlike previous methods that assume the identical distribution of information across channels, our approach considers the differences in noise type between channels. The bilateral filter, as shown in equation \ref{biteralgrid}, takes into account both the spatial distance and intensity difference between pixels, preserving edge information and enhancing image details and contrast. This helps prevent the generation of noise and artifacts in low-light scenes, leading to superior denoising results.

\begin{equation}
\hat{I}(p)=\frac{1}{W_{p}} \sum_{q \in S} I(q) f_{r}(\|p-q\|) f_{s}(\|I(p)-I(q)\|)
\label{biteralgrid}
\end{equation}
The bilateral filter is a spatial domain filter that is widely used in image processing tasks. It operates by calculating the weighted average of the pixels in a given neighborhood, where the weights depend on the distance between the pixels and their intensity differences. The filtered value of the pixel $p$ is denoted as $\hat{I}(p)$. The spatial domain of the filter is represented by $S$, while $W_p$ is the normalization factor. The range kernels for distance and intensity difference are denoted as $f_r$ and $f_s$, respectively. The range kernels determine the influence of a pixel $q$ on the filtered value of pixel $p$ based on the spatial distance and intensity difference between them. The greater the distance or intensity difference, the smaller the influence.

Bilateral grids \cite{chen2007bilateralgrid, chen2016bilateralgrid, bilateral} offer a way to extend bilateral filters to high-dimensional feature spaces, enabling the smoothing and enhancement of image features in a manner similar to conventional pixels. This is achieved by mapping image features onto a grid where the similarity between pixels is determined by the similarity between neighboring cells in the grid, based on spatial consistency.

To perform Deep Bilateral in this paper, the following steps are taken:
\begin{enumerate}
    \item The intermediate feature layer \textit{Fin} is used as input, and downsampling is applied to generate the bilateral grid with a size of (3,8,8,8).
    \item Trilinear interpolation is then used to obtain the coefficients (Coeffs) at the same resolution as the guidance map with dimensions of \textit{C*H*W}.
    \item The obtained coefficients are combined with the input feature map in the backbone, resulting in a single-channel filter.
    \item The combination operation is then performed separately on each of the three RGB channels to obtain the final denoised output.
\end{enumerate}

\subsection{Enhancement}
The enhancement part consists of a lightweight multi-branch restoration network that performs various restoration functions such as contrast enhancement, noise suppression, and color correction. Additionally, an illumination adjustment network is included to modify the light level.

\subsubsection{Restoration Net}

\begin{figure*}[htp]
    \centering
    \includegraphics[scale=0.6]{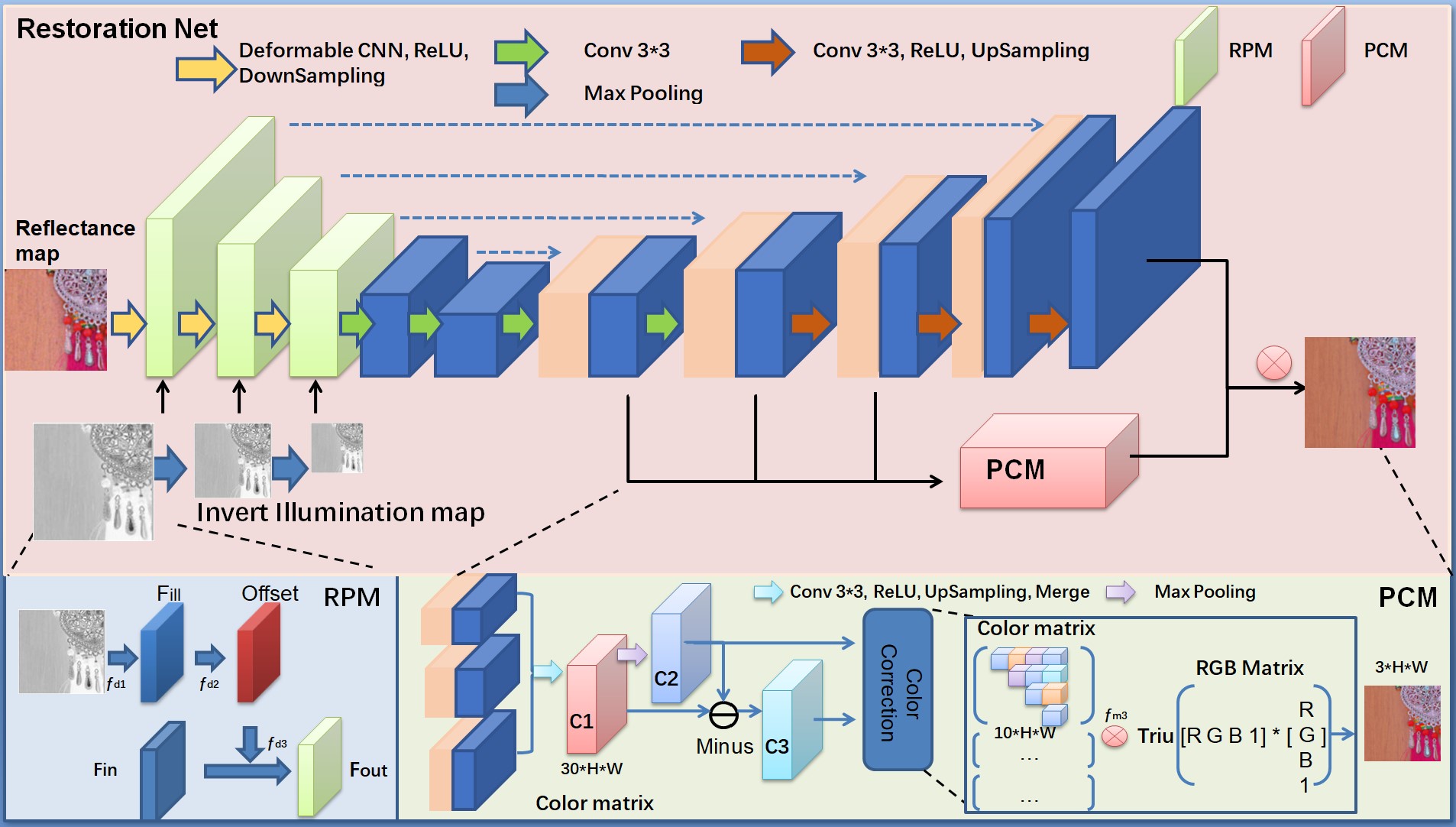}
    \caption{The architecture of Restorationnet. The encoder-decoder Net is divided into Regional Consistency-based Non-rigid Sampling Pyramid Module (RPM) and Progressive Binomial Color Correction  Module (PCM).}
    \label{fig:restore}
\end{figure*}

The Restoration net consists of an encoder-decoder framework and multiple branches designed for different restoration tasks, as shown in Figure \ref{fig:restore}. Specifically, it includes branches for detail restoration, denoising, and color correction.

\paragraph{Regional Consistency-based Non-rigid Sampling Pyramid Module (RPM)}

Images captured under low-light conditions or poor environmental settings often suffer from low contrast and blur, making it necessary to perform detail preservation to enhance the dark regions. To achieve this, we use the illumination image as a guide map to restore the details of the dark parts of the image in a targeted manner. To improve runtime efficiency, we invert the illumination map. In order to guide detailed recovery at different scales, we introduce a Laplace pyramid structure at three scales. This structure helps to preserve high-frequency details by processing information from the image at different scales, thereby avoiding the local artifacts that frequently occur in the output of optical enhancement networks when they are fed with high-resolution inputs. To implement the pyramid structure in our encoder, we use sequential maximum pooling to obtain a multi-scale illumination map that extracts hierarchical features.

The model aims to enhance the dark areas of the image, which are critical for overall improvement. Lighting is a spatially consistent feature that contributes to the region's overall appearance \cite{zou2007illumination, chen2018learning}. However, due to varying lighting conditions, the spatial distribution of dark areas may become irregular. Traditional convolutional operations may not effectively extract the relevant information, which is why deformable convolution \cite{dai2017deformable} is used instead. In contrast to other studies \cite{liang2020bilateralretinex,liu2023LAE}, we use the inverted illumination map as the offset map, directing the model's attention to the dark areas that contribute to noise and loss of detail. The offset map has a dimension of 2N, where N is the convolution kernel's area, providing the offset in the x and y directions. The adaptive sampling of the input feature map \textit{Fin} can then be performed according to the offset. To emphasize the dark areas as the key areas for detail recovery, we introduce an attention mechanism that allows the model to focus more on these areas, resulting in targeted detail recovery.

\paragraph{Progressive Binomial Color Correction  Module (PCM)}

In low light conditions, images often have significant color deviation due to the influence of noise in the camera sensor \cite{chen2018learning}. Therefore, color correction is necessary to ensure the accuracy and stability of the image.
To address this issue, we propose a novel approach that fully utilizes feature maps of different scales. Firstly, three feature maps of different scale sizes are adjusted to the same resolution, denoted as \textit{C1}, using convolution. Since color is spatially continuous and has a correlation between adjacent positions, maximum pooling is used to obtain local area information denoted as \textit{C2}, which considers information from multiple surrounding points and represents overall characteristics to some extent. However, the correction of \textit{C2} only yields a coarse result. To obtain finer correction information, fine-grained correction information \textit{C3} is obtained based on the offset between each point's own value (\textit{C1}) and the pooled value (\textit{C2}). Finally, \textit{C2} and \textit{C3} are superimposed to obtain the final color-corrected result.

Traditionally, previous studies \cite{bilateral,liang2020bilateralretinex} used a $3*3$ or $3*4$ matrix for color correction, which only allowed for RGB linear conversion with limited fitting effectiveness. In this paper, we improve upon this by utilizing a transformation method based on a $10*H*W$ color matrix, which is derived from the binomial expansion of the above-mentioned coefficients. The \textit{Triu (·)} function is used to vectorize the elements in the upper triangular matrix. This approach provides a better fit for nonlinear color correction, resulting in more accurate and stable color-corrected images.

\subsubsection{Illumination Adjustment Net}

To adjust the illumination of the input image, we used a network structure similar to the decomposition network, but with the CDM network removed. The illumination adjustment network is based on a combined architecture that includes a 6-layer UNET and a ResNet. This combined architecture utilizes both long-range and short-range connections to effectively create a correlation mapping that ensures both detail restoration and color reproduction in the output image.
Specifically, the UNET is responsible for capturing short-range correlations, while the ResNet is responsible for capturing long-range correlations. The UNET's skip connections effectively retain high-frequency information, which can be lost during the process of downsampling and upsampling. In contrast, the ResNet's residual connections allow for the effective propagation of information through multiple layers. 

\subsubsection{Loss Function}
The loss function used in this model is designed to optimize the performance of each individual network while also encouraging overall consistency between the different components.
\paragraph{Decomposition}
The decomposition loss $\mathcal{L}_{decom}$ is composed of two terms: The first term is the reflectance similarity, which measures the difference between the reflectance of the low and high images. The second term is  reconstruction loss, which measures the difference between the product of reflectance and illumination and the original RGB image for both the low and high images. The loss can be formulated as follows:
\begin{equation}
\mathcal{L}_{\text {decom}}=\|R_{\text {low }}-R_{\text {high }}\|_{1}+\sum_{i=\text { low,high }}\|R_{\text{i}} \circ I_{\text{i }} - S_{\text {i }}\|_{1} .
\end{equation}
where $R_{low}$ and $R_{high}$ denote the reflectance of the low and high images, $I$ and $S$ denote the illumination map and RGB image, and $\left|\right|_{1}$ denotes the $\zeta^{1}$ norm.

\paragraph{Restoration net}

The restoration loss $\mathcal{L}_{re}$ is composed of three terms. The first term is the reflectance similarity, which measures the difference between the restored reflectance and the reflectance of the high image. The second term is  reconstruction loss, which measures the difference between the restored reflectance and the high image using the structural similarity index (SSIM). The third term is edge similarity, which measures the difference between the first-order derivative of the restored reflectance and the high image. The loss can be formulated as follows:

\begin{equation}
\begin{aligned}
\mathcal{L}_{\text {re}}=\left\|\hat{\mathbf{R}}-\mathbf{R}_{high}\right\|_{2}^{2}-\operatorname{SSIM}\left(\hat{\mathbf{R}}, \mathbf{R}_{high}\right)
+\left\|\nabla \hat{\mathbf{R}}-\nabla \mathbf{R}_{high}\right\|_{2}^{2}
\end{aligned}
\end{equation}
where $\hat{R}$ corresponds to the restored reflectance, and SSIM(·, ·) is the structural similarity measurement. $\nabla$ stands for the first-order derivative operator containing $\nabla{x}$ (horizontal) and $\nabla{y}$ (vertical) directions. $\left|\right|_{2}$ denotes the $\zeta^{2}$ norm (MSE).
\paragraph{Illmination adjustment net}
The illumination adjustment loss $\mathcal{L}_{ill}$ is composed of two terms. The first term is the illumination map similarity, which measures the difference between the predicted illumination map and the high-image illumination map. The second term is edge similarity, which measures the difference between the first-order derivative of the predicted illumination map and the high-image illumination map. Formally, the loss can be formulated as follows:
\begin{equation}
\mathcal{L}_{\text {ill }}=\left\|\hat{\mathbf{I}}-\mathbf{I}_{high}\right\|_{2}^{2}+\left\|\nabla \hat{\mathbf{I}}-\nabla \mathbf{I}_{high}\right\|_{2}^{2}
\end{equation}

\section{Experimental Results}
In this section, we use a large number of experiments to evaluate our method. First, we compare our method with the current state-of-the-art enhancement methods in terms of quality and quantity. Then, we provide additional analyses to fully demonstrate the advantages of our method.

\subsection{Experimental Setups}
\paragraph{Training Data}
In our experiments, we employed two datasets for training our model: LOL \cite{retinex} and MIT5K \cite{MIT5K}. The LOL dataset contains 500 pairs of real-world low-light/normal-light images, which is the first dataset created for low-light image enhancement. We split the LOL dataset into three subsets with 400, 50, and 50 image pairs for training, validation, and testing, respectively. The MIT5K dataset contains 5,000 images, out of which we used 4,500 images for training, and the remaining 500 images were used for validation and testing.

\paragraph{Evaluation metrics} We used two widely adopted metrics, PSNR (Peak Signal to Noise Ratio) and SSIM (Structural Similarity Index), to evaluate the quality of the enhanced images compared to the ground truth images. A higher PSNR value indicates less distortion between the two images, while a higher SSIM value indicates higher structural similarity between the two images. Both metrics are commonly used in image processing to measure the quality of image enhancement. In general, higher PSNR and SSIM values indicate better results and a more realistic human perception of the images.

\paragraph{Implementation Details}

We implemented our proposed method using the PyTorch framework and trained it on an Nvidia 2080T GPU. To prevent overfitting and enhance the generalization ability of the model, randomly, flipped, and rotated the images during training. We used the Adam optimizer \cite{ADAM} to optimize the network, with hyperparameters set as follows: $\alpha$ = 0.001, $\beta_1$ = 0.9, $\beta_2$ = 0.999, and
$\epsilon$ = $10^{-8}$. 

\begin{figure*}[tb]
    \centering
  \subfigure[Input\label{input}]{%
       \includegraphics[width=0.25\linewidth]{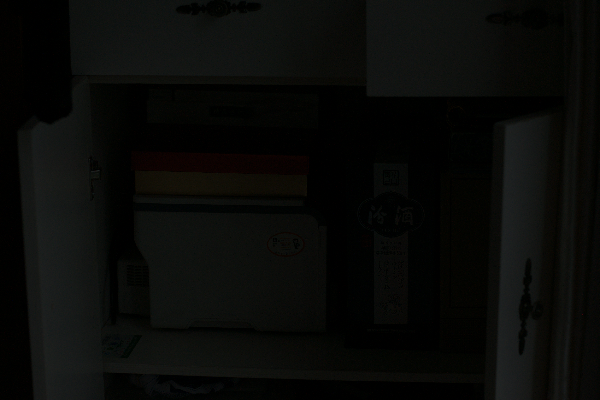}}
    \subfigure[GT\label{GT}]{          
        \includegraphics[width=0.25\linewidth]{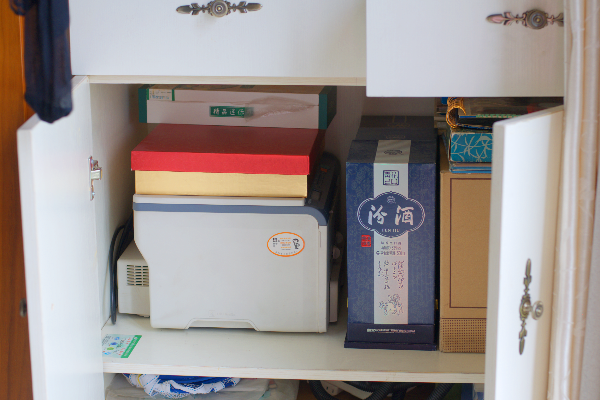}}
  \subfigure[RetinexNet\label{retinex}]{%
        \includegraphics[width=0.25\linewidth]{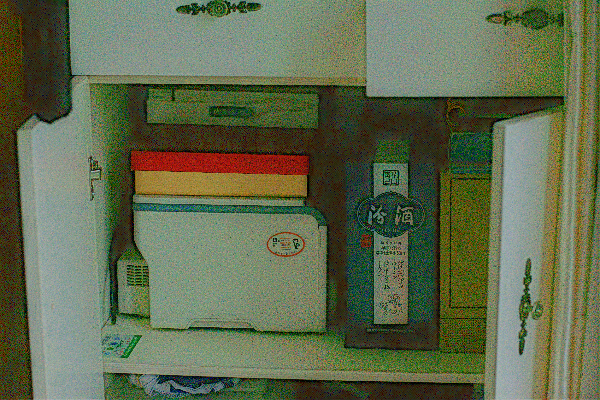}}
        \\
  \subfigure[KinD\label{kind}]{%
        \includegraphics[width=0.25\linewidth]{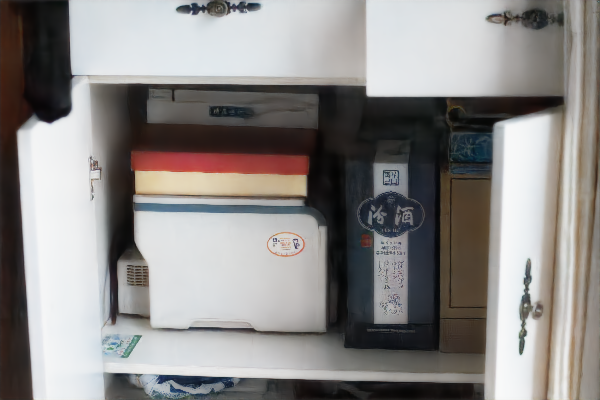}}
  \subfigure[KinD++\label{kind_plus}]{%
        \includegraphics[width=0.25\linewidth]{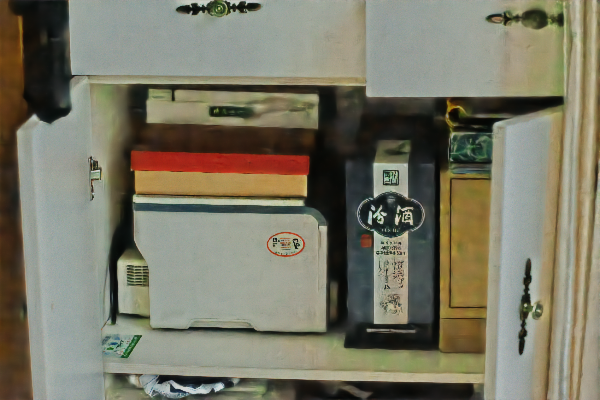}}
  \subfigure[ZeroDCE\label{zero_DCE}]{%
        \includegraphics[width=0.25\linewidth]{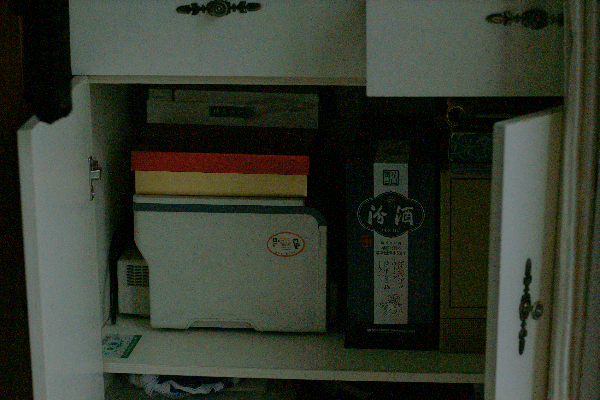}}
    \\
  \subfigure[DeepUPE\label{Deep_UPE}]{%
        \includegraphics[width=0.25\linewidth]{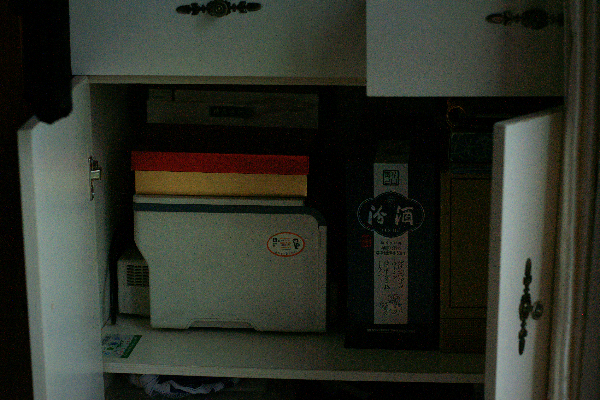}}
  \subfigure[IAT\label{IAT}]{%
        \includegraphics[width=0.25\linewidth]{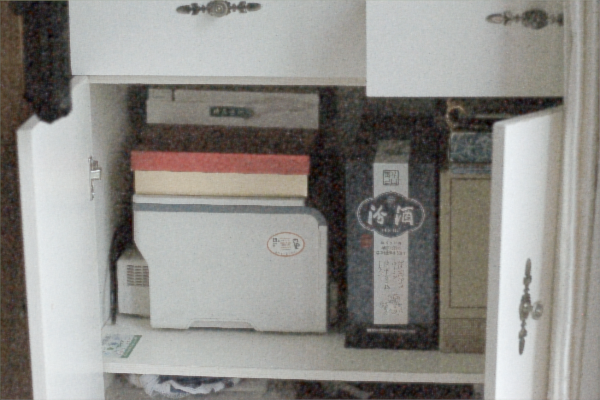}}
  \subfigure[Ours\label{ours}]{          
    \includegraphics[width=0.25\linewidth]{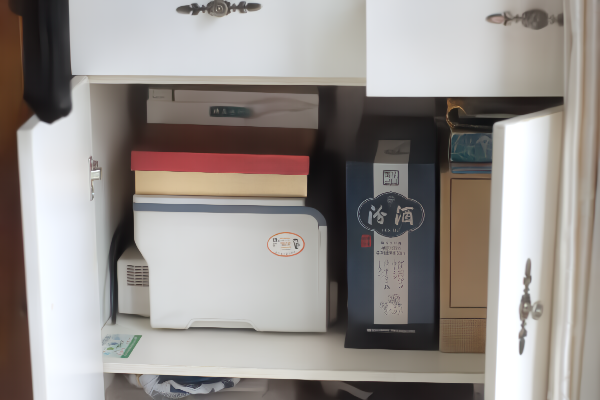}}

  \caption{Visual results on LOL dataset}
  \label{fig:visual} 
\end{figure*}

\subsection{Qualitative Results}
We compared the performance of our proposed method with state-of-the-art enhancement methods in terms of both quality and quantity. We also conducted additional analyses to demonstrate the advantages of our method. Specifically, we compared our proposed method with RetinexNet \cite{retinex}, KinD \cite{kind}, KinD++ \cite{kind_plus}, DeepUPE \cite{DeepUPE}, and ZeroDCE \cite{ZeroDCE}, on two publicly available datasets: LOL and MIT5K. Our model was trained on the LOL dataset and evaluated on both datasets using two commonly used image quality metrics, namely PSNR and SSIM \cite{SSIM}. Table \ref{tab:lol} reports the quantitative comparison results, where our proposed method achieves superior performance in terms of both metrics.
 
 Based on Table \ref{tab:lol}, our proposed method demonstrates superior performance compared to existing state-of-the-art methods in terms of PSNR and SSIM on the LOL dataset. Our method achieves the best performance with an average PSNR of 23.162 dB and SSIM of 0.835, which is 0.481 dB higher than the second-best method (IAT) in PSNR and 0.017 in SSIM. It is also 1.851 dB higher than the third-best method (KinD++) in PSNR and 0.024 in SSIM. However, RetinexNet, DeepUPE, and ZeroDCE are not as effective. On the other hand, as shown in Table \ref{tab:mit5k}, our proposed method achieves the highest scores with PSNR of 26.16 dB and SSIM of 0.962 on the MIT5K dataset.

\begin{table}[tb]
\centering
\begin{tabular}{|c|c|c|c|}
\hline
Datasets             & Method        & PSNR            & SSIM           \\ \hline
LOL & Retinex-Net   & 16.674          & 0.490          \\ \cline{2-4} 
                     & KinD          & 20.882          & 0.791          \\ \cline{2-4} 
                     & KinD++        & 21.311          & 0.821          \\ \cline{2-4} 
                     & ZeroDCE      & 14.96           & 0.573          \\ \cline{2-4} 
                     & DeepUPE       & 16.798          & 0.519          \\ \cline{2-4} 
                     & IAT           & 22.681          & 0.818          \\ \cline{2-4} 
                     & \textbf{ours} & \textbf{23.162} & \textbf{0.835} \\ \hline
\end{tabular}
\caption{Quantitative comparison on LOL dataset in terms of PSNR, SSIME. The best results are highlighted in bold.}
\label{tab:lol}
\end{table}

\begin{table}[tb]
\centering
\begin{tabular}{|c|c|c|c|}
\hline
Datasets               & Method        & PSNR           & SSIM           \\ \hline
MIT5K & RetinexNet   & 14.80          & 0.720          \\ \cline{2-4} 
                       & KinD          & 17.58          & 0.686          \\ \cline{2-4} 
                       & DeepUPE       & 23.04          & 0.893          \\ \cline{2-4} 
                       & ZeroDCE      & 16.99          & 0.813          \\ \cline{2-4} 
                       & IAT           & 25.32          & 0.920          \\ \cline{2-4} 
                       & \textbf{ours} & \textbf{26.16} & \textbf{0.962} \\ \hline
\end{tabular}
\caption{Quantitative comparison on MIT5K dataset in terms of PSNR, SSIME. The best results are highlighted in bold.}
\label{tab:mit5k}
\end{table}

\subsection{Visual Comparisons}

 We conducted a visual comparison of the enhancement results produced by our proposed method and other Retinex theory-based methods (e.g. KinD, RetinexNet) as shown in Figure \ref{fig:visual}. The results indicate that our method effectively enhances the contrast, improves details, and removes noise, while previous methods tend to blur the details or amplify the noise. Specifically, RetinexNet produces images with significant color distortion and noise, while KinD++ results are slightly dark and lack subtle details. ZeroDCE and DeepUPE produce dimmer images with heavily hidden details, and IAT improves brightness but still suffers from noise and color distortion. In contrast, our method recovers true colors and obtains more texture details without noise compared to other methods.

In general, our method is capable of enhancing dark areas without introducing overexposure artifacts and maintaining high-contrast texture details. This can be attributed to the pyramid structure, luminance-aware guidance, and contrast-attentive mechanism, which all contribute to the model's ability to predict reasonable adjustments and reconstruct high-quality images.
\subsection{Ablation Study}
\begin{figure}
\centering
  \subfigure[Baseline\label{input}]{%
       \includegraphics[width=0.25\linewidth]{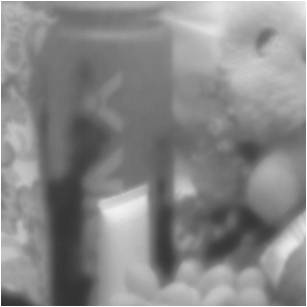}}
    \subfigure[w/CFM\&CDM\label{GT}]{        
        \includegraphics[width=0.25\linewidth]{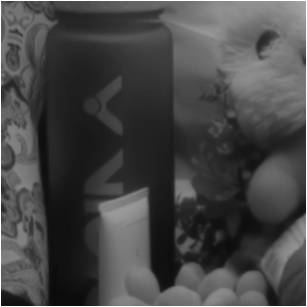}}
  \subfigure[GT\label{retinex}]{%
        \includegraphics[width=0.25\linewidth]{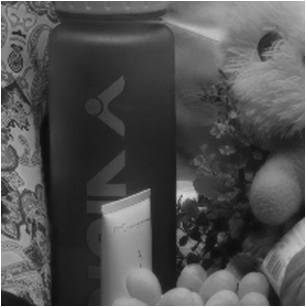}}
  \caption{Illumination maps of baseline (a), with CFM and CDM (b) from low-light images.  (c) shows a illumination map with CFM and CDM from normal-light images.}
  \label{fig:ill} 
\end{figure}

\begin{figure}[tb]
    \centering
  \subfigure[Baseline\label{input}]{%
       \includegraphics[width=0.25\linewidth]{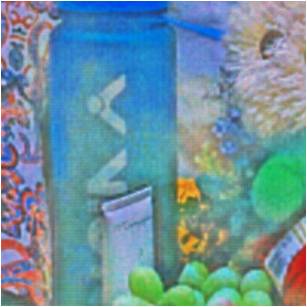}}
    \subfigure[w/CFM\&CDM\label{GT}]{        
        \includegraphics[width=0.25\linewidth]{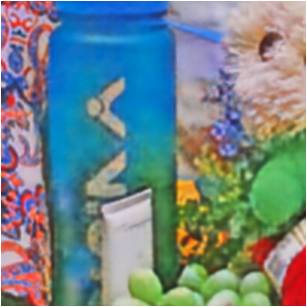}}
  \subfigure[w/all modules\label{retinex}]{%
        \includegraphics[width=0.25\linewidth]{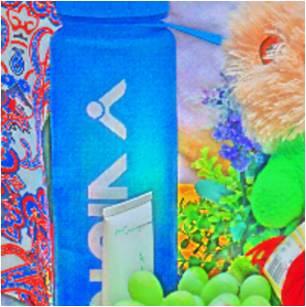}}
  \subfigure[GT\label{retinex}]{%
\includegraphics[width=0.25\linewidth]{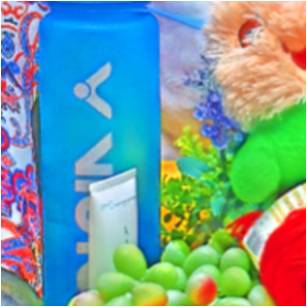}}
  \caption{Reflectance maps of baseline (a), with CFM and CDM (b), with all modules (c) from low-light images.  (d) shows a reflectance map with all modules from normal-light images.}
  \label{fig:reflectace} 
\end{figure}

We also conducted an ablation study on the LOL dataset by removing each of the four modules (CFM, CDM, RPM, PCM) individually to assess the effectiveness of each module in our network. Table \ref{ablation} shows the results, indicating that removing the CDM and PCM modules leads to a significant decrease in the performance of our model. For more detailed analysis, please refer to Figures \ref{fig:ill} and \ref{fig:reflectace}.

\begin{table}[]
\centering
\begin{tabular}{llllll}
\hline
Case & CFM & CDM & RPM & PCM & LOL    \\ \hline
1    &  & $\checkmark$  & $\checkmark$  & $\checkmark$  & 22.925 \\
2    & $\checkmark$  &  & $\checkmark$  & $\checkmark$  & 21.964 \\
3    & $\checkmark$  & $\checkmark$  &  & $\checkmark$  & 22.647 \\
4    & $\checkmark$  & $\checkmark$  & $\checkmark$  &  & 21.683 \\
5    & $\checkmark$  & $\checkmark$  & $\checkmark$  & $\checkmark$  & 23.162 \\ \hline
\end{tabular}
\caption{Ablation of CFM, CDM, RPM and PCM on LOL dataset in terms of PSNR. }
\label{ablation}
\end{table}

\section{Conclusion}
In this paper, we propose a new end-to-end enhancement network based on Retinex theory for low-light pictures and normal-light pictures. The network consists of two parts: the decomposition network, and the enhancement network. In this paper, guided by spatial consistency, we combine semantic and texture information, bilateral filtering, and binomial color correction to edge recovery, denoise, and color correction for a single channel of RGB image, so that the enhancement results obtained by our method have better visual effects. And it is verified that our scheme of low illumination image recovery is correct and feasible. Experimental results on the LOL dataset show that our method can improve the image contrast and good noise rejection, and obtain the highest PSNR and SSIM scores, which are superior to other methods.

\bibliographystyle{unsrt}  
\bibliography{references}  
\end{document}